# MLLM-Search: A Zero-Shot Approach to Finding People using Multimodal Large Language Models

Angus Fung, Aaron Hao Tan, Haitong Wang, *Student Member, IEEE,* Beno Benhabib, and Goldie Nejat, *Member, IEEE*

*Abstract*—Robotic search of people in human-centered environments, including healthcare settings, is challenging as autonomous robots need to locate people without complete or any prior knowledge of their schedules, plans or locations. Furthermore, robots need to be able to adapt to real-time events that can influence a person's plan in an environment. In this paper, we present MLLM-Search, a novel zero-shot person search architecture that leverages multimodal large language models (MLLM) to address the mobile robot problem of searching for a person under event-driven scenarios with varying user schedules. Our approach introduces a novel visual prompting method to provide robots with spatial understanding of the environment by generating a spatially grounded waypoint map, representing navigable waypoints by a topological graph and regions by semantic labels. This is incorporated into a MLLM with a region planner that selects the next search region based on the semantic relevance to the search scenario, and a waypoint planner which generates a search path by considering the semantically relevant objects and the local spatial context through our unique spatial chain-of-thought prompting approach. Extensive 3D photorealistic experiments were conducted to validate the performance of MLLM-Search in searching for a person with a changing schedule in different environments. An ablation study was also conducted to validate the main design choices of MLLM-Search. Furthermore, a comparison study with state-of-the art search methods demonstrated that MLLM-Search outperforms existing methods with respect to search efficiency. Real-world experiments with a mobile robot in a multi-room floor of a building showed that MLLM-Search was able to generalize to finding a person in a new unseen environment.

*Index Terms*— Robotic Person Search, Multimodal Large Language Models, Zero-Shot Search, Event-driven Scenarios

## I. INTRODUCTION

Autonomous mobile robots can be used to search for specific people in human-centered environments to engage in human-robot interactions. For example, robots need to locate people in: 1) multi-room homes to assist with daily tasks such as meal preparation and exercise [1]–[3], 2) office and university buildings to deliver packages or messages [4]–[6], and 3) public venues such as shopping malls and amusement parks to locate lost people [7]–[11]. In healthcare settings, robots search in long-term homes to find residents to remind them of meal-times or appointments [12], [13] and in hospitals to find medical professionals to deliver supplies [4], [14], [15], or to guide visitors to their destinations [16].

Existing person search methods such as Hidden Markov Model (HMM)-based [3]–[5], [17] and Markov Decision Process (MDP)-based [18]–[21] planners have been used by robots to search for individuals with known user models. These models are generated from user location patterns, such as daily schedules [1], [3], [19], [21] or past locations [17], [18]. However, user schedules may be unavailable or incomplete, especially for new users, and may change unexpectedly due to real-time events (*e.g.*, weather, delays in appointments). As these methods rely on past user behavior patterns, they are unable to generalize to new scenarios.

Recently, multimodal large language models (MLLMs) [22]–[25] have been proposed for robotic tasks such as robot navigation to unknown static objects. As MLLMs are trained on extensive data obtained from the internet [26], they have generalist reasoning capabilities [27]. Furthermore, in search tasks within known environments where metric maps are available, MLLMs, such as GPT-4o [28], can directly interpret the spatial layouts of an environment from the metric maps for search planning. However, MLLMs have limited spatial reasoning capabilities due to being trained on image-caption pairs, which contain limited spatial information [29], [30]. To address this, visual prompting methods have been developed [31]–[34]. These methods overlay coordinates representing locations of objects for general visual Q&A tasks [31]–[33] or robot actions for navigation tasks [34] directly onto RGB images to improve spatial understanding of the local scene. However, these visual prompting methods cannot be directly applied to the robotic person search problem, as they lack the spatial understanding of the overall environment needed when searching within a known environment.

MLLMs have the potential to be applied to robotic person search problems by leveraging their reasoning capabilities [27] to infer the location of a dynamic individual from incomplete schedules. Moreover, they can incorporate new/additional information within the MLLM's context window, enabling zero-shot person search without retraining.

In this paper, we present a novel multimodal language model, *MLLM-Search*, to address for the first time the robotic person search problem under event-driven scenarios where user schedules are incomplete, unavailable or deviate due to real-time events. The main contributions of this paper are:
1) the development of the first zero-shot person search method which incorporates language models for generalist reasoning and spatial understanding of the environment; 2) the introduction of a novel visual prompting method that generates a topological graph with semantic region labels by extracting spatial information from metric maps. The novelty lies in generating a semantically and spatially grounded waypoint map that uniquely enables MLLMs to perform search planning by providing spatial reasoning of the overall global environment; and 3) the development of MLLM-based

This research is supported by the Natural Sciences and Engineering Research Council of Canada (NSERC) and the Canada Research Chairs (CRC) program. The authors are with the Autonomous Systems and Biomechatronics Lab in the Department of Mechanical and Industrial Engineering at the University of Toronto, 5 King's College Road, Toronto, ON, M5S 3G8 Canada. (Email: {angus.fung, haitong.wang}@mail.utoronto.ca, aaronhao.tan@utoronto.ca, {benhabib, nejat}@mie.utoronto.ca).

search planner which incorporates a region planner and a waypoint planner. The region planner considers the semantic relevance of each search region with respect to the event-driven search scenario using region-based score prompting. The waypoint planner uses semantically relevant objects in the environment during planning within our new spatial chain-of-thought (SCoT) prompting method. Our search planner is able to generalize to scenarios with varying user schedules where historical data is unavailable to optimize the likelihood of finding a person of interest.

## II. RELATED WORKS

In this section, we present and discuss existing: 1) person search methods developed for robotic search of a dynamic person in human-centered environments, and 2) visual prompting methods used to improve spatial reasoning.

### A. Person Search by Robots

Existing person search methods for robotic applications have consisted of: 1) lookahead [3]–[5], [17], 2) MDP-based [18]–[21], or 3) graph-based [1], [2] planners.

*1) Lookahead Planners*

Lookahead planners have either used HMMs [3], [4] or predefined likelihood functions [5], [17] to navigate a robot to the most probable user locations. In particular, HMMs identify user locations based on past location and activity data. For example, in [3], the Casper robot used an HMM to find static known people in a single-floor home. The HMM predicted a person's location based on their past activity sequence. In [4], an HMM-based person search method was used by a robot to find a dynamic person in an indoor office setting. The HMM predicted the person's movements in the environment based on past observed locations.

In [5] and [17], lookahead planners used likelihood functions defined by human-experts [5] or by past user locations [17] to determine the next search region. Namely, in [5], a robot searched for static people in an indoor laboratory, navigating to locations with the highest likelihoods assigned by human-experts. In [17], a robot searched for a dynamic resident in an apartment by navigating to the highest likelihood location based on past user location frequencies.

*2) MDP-based Planners*

MDP-based planners [18]–[21] have optimized robotic search actions by maximizing the likelihood of finding a person. For example, in [18], an MDP-based method was used by a robot to search for static people on a floor of a building. A sequence of search regions was determined based on the expected proportion of people in each region to minimize search time. In [20], an MDP-based method was also used for a mobile robot to find an elderly person in a home. The MDP determined the next location to visit based on the probability of the person's current location. In [21], an MDP search method generated a search plan to maximize the number of residents found within a retirement home in a specific time frame. The search plan consisted of actions such as moving to different regions. Likelihood functions, based on residents' daily activity schedules, were used to compute the reward.

In [19], a partially observable MDP approach was used for a robot to find a dynamic person in a multi-room home environment. The approach determined search actions (*e.g.*, searching a room) by incorporating user activity data (*e.g.*, meal preparation) into a Bayesian network to determine the highest likelihood user locations.

*3) Graph-based Planners*

Graph-based planners have used activity probability density functions (APDF) to plan search paths based on user schedules. For example, in [1], a people search method was proposed for the assistive robot Blueberry to search for dynamic people in long-term care homes. APDF was utilized to predict user locations by considering their complete schedules. These schedules included activity types and duration, time of day, and specific regions. The work in [1] was extended in [2] to consider multiple robots searching.

### B. Visual Prompting Methods

Visual prompting methods [31]–[34] for MLLMs improve spatial reasoning over standard prompting methods by overlaying visual coordinates onto RGB images of a scene. In [31]–[33], visual prompting methods were introduced for visual Q&A tasks to infer object positions [31], to identify regions based on text [32], or to answer queries related to size and distance of objects in a scene [33]. For example, in [31], the Scaffold method placed visual coordinates evenly across an RGB image of a scene, allowing MLLMs to associate visual data with textual data. Similarly, both [32] and [33], placed visual coordinates on segmented objects within an RGB image of a scene to associate these objects with their corresponding visual coordinates. Visual prompting methods have also been used in robot manipulation [34]. In [34], the PIVOT method applied visual prompting to robot manipulation tasks by placing coordinates within an RGB image of a scene corresponding to potential robot manipulation actions. The MLLM was then used to select the next robot action based on the coordinates from the image.

### C. Summary of Limitations

Existing lookahead planners [3]–[5], [17] prioritize the next user region to search with the highest likelihood. However, this can result in increased search times as search plans may select further away regions [1]. On the other hand, MDP-based planners [18]–[21] rely on the Markov assumption that search decisions are based solely on the current region. This can result in redundant searches, where a robot revisits a recently searched region. Graph-based planners along with MDP-based and lookahead planners, all require complete user schedules [1], [3], [19], [21], past observed user locations [17], [18], and/or last known user locations [20]. However, in real-world scenarios, such user information may be unavailable or incomplete due to insufficient knowledge about the user. Furthermore, user behaviors can deviate from expected schedules due to real-time events such as emergency situations or schedule changes, availability of locations, etc. Thus, existing robotic person search methods cannot generalize to these real-world scenarios.

MLLMs have the potential to infer the region a person is in from incomplete or changing information, by leveraging knowledge learned from extensive internet data [26]. They can also consider additional search information beyond user schedules, such as building/room schedules, or activity-specific data, without requiring retraining or specialized models for each data type. However, MLLMs have not yet been applied to robotic person search tasks. While existing MLLMs have incorporated visual prompting methods [31]–[34], these methods have focused on improving spatial understanding of local scenes from RGB images. However, spatial understanding of the entire environment is needed when searching for individuals within known environments.

To address the aforementioned limitations, we propose *MLLM-Search*, the first robotic person search method that leverages MLLMs for event-driven scenarios where user schedules are varying. *MLLM-Search* provides spatial reasoning of the global environment by generating a semantically and spatially grounded waypoint map. Our search method incorporates both region and waypoint planners to generalize to scenarios with user schedules that are varying in completeness or have changed.

## III. PERSON SEARCH PROBLEM UNDER EVENT-DRIVEN SCENARIOS WITH VARYING USER SCHEDULES

### A. Problem Definition

The robot problem of person search under event-driven scenarios requires a mobile robot to search for a dynamic person in a known environment without complete, partial, or any *a priori* knowledge of their schedules. A search query $q_s$, provided by the search operator to the robot, includes natural language instructions containing the person to search for and their physical description $q_a$. The search query can optionally contain information such as their name, role, tasks, last known location, *etc*. The robot has access to an information database $Q_{db}$ which consists of: 1) the user schedule $Q_u$, if available, and 2) the system database $Q_s$, if available, containing textual data relevant to the search, such as building/room schedules, visitor logs, EHRs (Electronic Health Records), *etc*. During the search at time $t$, images $\mathbf{x}_t$ are obtained from the robot's camera. The function $f$ is defined to output a sequence of robot actions $\mathbf{u}_t$ given the robot position $\mathbf{p}_i$, metric map $\mathcal{M}$, search query $q_s$, and information database $Q_{db}$:
$$u_t = f(\mathbf{x}_t, \mathbf{p}_i, \mathcal{M}, q_s, Q_{db}). \quad (1)$$
The objective is to minimize the expected distance traveled $d$ between the robot start location $\mathbf{p}_s$ and the target location $\mathbf{p}_{tg}$:
$$\min \mathbb{E}\left[d(\mathbf{p}_s, \mathbf{p}_{tg})\right]. \quad (2)$$

## IV. MLLM-SEARCH ARCHITECTURE

The proposed *MLLM-Search* architecture, Fig. 1, consists of two subsystems: 1) *Map Generation Subsystem (MGS)*, and 2) *Person Search Subsystem (PSS)*. The goal of the *MGS* is to generate both a semantic metric map $\mathcal{M}_{sem}$ and waypoint metric map $\mathcal{M}_{wp}$ of the environment. Once the environment is mapped, the *PSS* leverages MLLMs to search for the user.

### A. Map Generation Subsystem (MGS)

The *MGS* consists of two main modules: the *Semantic Map Generation (SMG)* module and the *Waypoint Map Generation (WMG)* module.

*1) Semantic Map Generation (SMG)*

The *SMG* module builds a semantic map of the environment. It consists of three modules: 1) *Object Discovery VLM (OD-VLM)*, 2) *Open Segmentation (OS)*, and 3) *Semantic Simultaneous Localization and Mapping (S-SLAM)*. The *OD-VLM* module utilizes an MLLM to identify the objects in the environment, Namely, it takes an input image $\mathbf{x}_{RGB}$ and generates a list of detected objects labels $\mathbf{L}_o$ in the image. The *OS* module takes these object labels $\mathbf{L}_o$ and uses the Grounded Segment Anything Model (Grounded SAM) [35] to generate corresponding segmentation masks $\mathcal{M}_{seg}$ for each object. The *S-SLAM* module [36] takes an RGB image $\mathbf{x}_{RGB}$ and depth image $\mathbf{x}_D$, and produces a semantic map $\mathcal{M}_{sem}$, Fig. 2(a). Specifically, it takes the segmented portions of $\mathbf{x}_{RGB}$ and $\mathbf{x}_D$, and projects them into a 3D point cloud $\mathbf{x}_{SEG-PCL}$ using the pinhole camera model [37], [38]. The point cloud $\mathbf{x}_{SEG-PCL}$ is then converted into a voxel representation and summed over the height dimension to obtain the semantic map [36]. The semantic map is updated during the search, to represent new objects and existing objects that have changed locations.

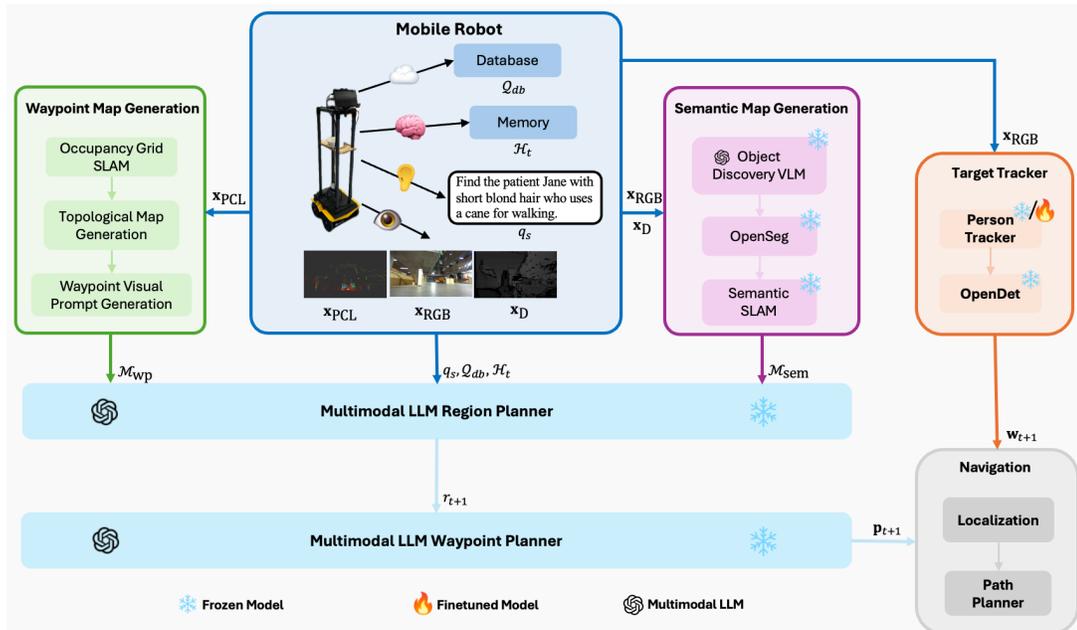

Fig. 1: Proposed *MLLM-Search* architecture for person search.

and region labels $L_r$ are represented as text (e.g., "Main Lobby"), Fig. 2(b). The map $\mathcal{M}_{wp}$ is passed into the *PSS*.

### B. Person Search Subsystem

The *Person Search Subsystem* is used to search for individuals within a dynamic environment using the semantic and waypoint maps generated by *MGS*. It contains the following modules: *MLLM Region Planner*, *MLLM Waypoint Planner*, the *Target Tracking*, and *Navigation*.

*1) Multimodal LLM Region Planner*

The *MLLM Region Planner* ($MLLM_{RP}$) module determines the region $r_{t+1} \in \mathcal{R}$ to search by considering the semantic relevance of each search region with respect to the search scenario. The inputs to this module include the robot's waypoint position $\mathbf{w}_t$, the search query $q_s$, the information database $Q_{db}$, and the robot memory $\mathcal{H}$. $\mathcal{H}$ consists of previous search histories for regions visited $\mathcal{H}_r = \{r_i\}_{i=1}^t$ and waypoints visited $\mathcal{H}_w = \{\mathbf{w}_i\}_{i=1}^t$, namely, $\mathcal{H} = \mathcal{H}_r \cup \mathcal{H}_w$. Region-to-object assignments are obtained from the semantic map $\mathcal{M}_{sem}$ by assigning each object to the closest region: $\mathcal{O}_r = \{(r_i, \{o_j\}_{j \in J_i})\}_i$. $J_i$ is the index set of objects $o_j$ assigned to region $r_i$. The contextual database $Q'_{db}$, representing semantically relevant information for the search, is retrieved through Retrieval Augmented Generation (RAG) [41]. RAG uses the cosine similarity between the search query embeddings $e(q_s)$ and database embeddings $e(q_{db}^i)$, where the database is divided into chunks $q_{db}^i \in Q_{db}$:

$$Q'_{db} = \arg\max_{Q_{db}} \frac{e(q_s) \cdot e(q_{db}^i)}{\|e(q_s)\| \|e(q_{db}^i)\|}. \quad (5)$$

We use region-based score prompting to assign semantic scores, $S_{t+1} = \{(s_l^i, s_p^i, s_r^i)\}$, to potential search regions $r_i$. $MLLM_{RP}$ outputs the semantic scores for each region $r_i$:

$$S_{t+1} = MLLM_{RP}(q_s, Q'_{db}, \mathbf{w}_t, \mathcal{H}_t, \mathcal{O}_r, \mathcal{M}_{wp}). \quad (6)$$

In particular, $S_{t+1}$ consists of: 1) the likelihood score $s_l$, 2) the proximity score $s_p$, and 3) the recency score $s_r$ shown in Fig. 3. Furthermore, we use Chain-of-Thought (CoT) prompting [42] to provide explicit and sequential step-by-step reasoning. The text and visual prompts are also presented in Fig. 3. The robot selects the region with the highest sum of semantic scores as the next region $r_{t+1}$ to search:

$$r_{t+1} = \arg\max_{\mathcal{R}} s_l^i + s_p^i + s_r^i. \quad (7)$$

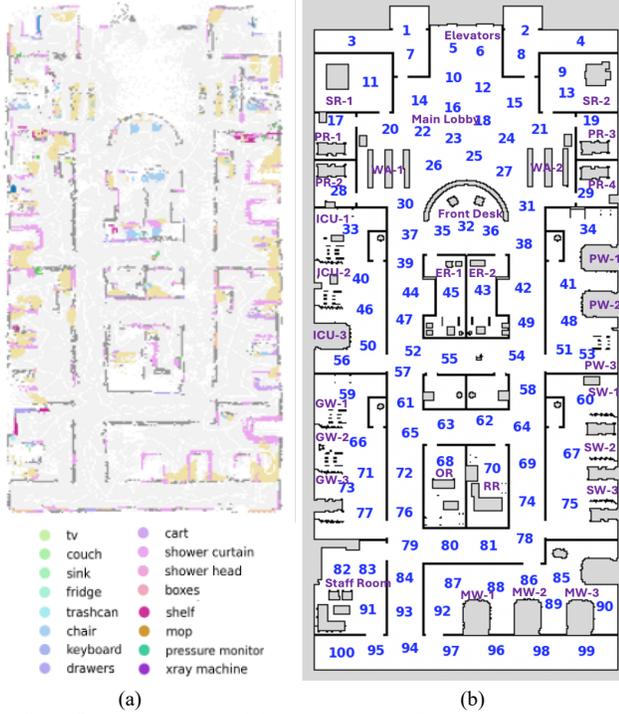

(a)     (b)

Fig. 2: (a) Semantic map, and (b) Waypoint map of a hospital environment.

*2) Waypoint Map Generation (WMG)*

The *WMG* module generates a semantically and spatially grounded waypoint map $\mathcal{M}_{wp}$ and consists of three sub-modules: 1) *Occupancy Grid SLAM (OG-SLAM)*, 2) *Topological Map Generation (TMG)*, and 3) *Waypoint Visual Prompt Generation (WVPG)*. The *OG-SLAM* sub-module creates an occupancy map $\mathcal{M}_{occ}$ using odometry $\rho$ and point clouds $\mathbf{x}_{PCL}$ with particle filters [39]. The *TMG* sub-module uses the occupancy map $\mathcal{M}_{occ}$ to generate a topological map, represented as a graph $\mathcal{M}_{top} = (V, E)$, where nodes $V$ represent navigable waypoints, and edges $E$ represent the traversable paths between waypoints. The nodes $V$ are obtained based on free space in $\mathcal{M}_{occ}$. First, the distance transform $D$ is computed, which measures the distance from each point $\mathbf{p}$ on $\mathcal{M}_{occ}$ to the nearest obstacle $\mathbf{o} \in O$. For each point $\mathbf{p}$ in $\mathcal{M}_{occ}$, the distance transform is computed:

$$D(\mathbf{p}) = \min_{\mathbf{o} \in O} |\mathbf{p} - \mathbf{o}|. \quad (3)$$

Safe points $\mathbf{p}_{safe} = \{\mathbf{p} \mid D(\mathbf{p}) \geq \sigma_{min}\}$ are identified as those a distance $\sigma_{min}$ away from obstacles. K-means clustering is used to generate waypoints $\mathbf{w}_i$ from these safe points. Any $\mathbf{w}_i$ such that $D(\mathbf{w}_i) < \sigma_{min}$ are updated to the nearest waypoint:

$$\mathbf{w}_i' = \arg\min_{\mathbf{p} \in S} |\mathbf{p} - \mathbf{w}_i|. \quad (4)$$

Edges $E$ are determined using a KDTree to find neighboring nodes within a distance $\sigma_{max}$. Bresenham's algorithm [40] is used to check if the path between waypoints is obstacle-free.

The *WVPG* sub-module uses both $\mathcal{M}_{occ}$ and $\mathcal{M}_{top}$ to generate a waypoint map $\mathcal{M}_{wp}$, Fig. 2(b), where navigation waypoints $\mathbf{w}_i$ and high-level region labels $L_r$ are directly overlaid on top of the occupancy map. Waypoints $\mathbf{w}_i$ are represented as numbered markers (i.e., $\mathbf{w}_1$ is labelled as "1"),

**Visual Prompt:** Waypoint map with semantic room labels: $\mathcal{M}_{wp}$
**Textual Prompt:** You are a robot tasked to locate a person by generating a search plan. The current time is $t$, and your current location is at $\mathbf{w}_t$.

Search query: $q_s$, $Q'_{db}$.
Search history: $\mathcal{H}$. Region-to-object assignments $\mathcal{O}_r$.
Your search strategy should include:
1. A likelihood score representing the likelihood of the person being present in each room.
2. A proximity score assigned to each room based on its distance from your current location, with priority given to nearest rooms.
3. A recency score which reduces the score for rooms that have been recently searched to minimize repeat searches.

Provide a complete analysis of each region in the following JSON format: {"region": "<region_name>", "likelihood": <score>, "proximity" : <score>, "recency": <score> "reason": "<step_by_step_explanation>"}.

Fig. 3: Text and visual prompt of the MLLM Region Planner

*2) Multimodal LLM Waypoint Planner*

The *MLLM Waypoint Planner* ($MLLM_{WP}$) module plans a sequence of waypoints $\mathbf{p}_{t+1}$ to a search region $r_{t+1}$, while prioritizing the likelihood of encountering the person along the path. For example, when searching for a student, it may plan a route near tables where students work. A* [43] is used to generate several paths $\mathbf{p}_i$ from the current waypoint $\mathbf{w}_t$ to the destination waypoint $\mathbf{w}_{t+1}$. Waypoint-to-object assignments are obtained from $\mathcal{M}_{sem}$ by associating each object to the closest waypoint: $\mathcal{O}_w = \{(\mathbf{w}_i, \{o_j\}_{j \in J_i})\}_i$. This allows $MLLM_{WP}$ to consider semantically relevant objects during planning. We have developed a novel spatial CoT (SCoT) prompting method which improves spatial awareness of MLLMs over standard prompting methods by decomposing the planning task into sequential steps, with each step uniquely considering the semantically relevant objects (the parameter "objects" in Fig. 4) as well as the local spatial context (the parameter "next_waypoints" in Fig. 4). $MLLM_{WP}$ finds $\mathbf{p}_{t+1}$ by optimizing the path to region $r_{t+1}$:

$$\mathbf{p}_{t+1} = MLLM_{WP}(q_s, Q'_{db}, \mathbf{w}_t, r_{t+1}, \mathbf{p}_i, \mathcal{H}_t, \mathcal{M}_{wp}, \mathcal{O}_w). \quad (8)$$

The output waypoint sequence $\mathbf{p}_{t+1}$ is checked for feasibility against the topological map $\mathcal{M}_{top}$:

$$\forall (\mathbf{w}_i, \mathbf{w}_{i+1}) \in \mathbf{p}_{t+1}, (\mathbf{w}_i, \mathbf{w}_{i+1}) \in \mathcal{M}_{top}. \quad (9)$$

If constraints are violated, the above steps are repeated.

> **Visual Prompt:** Waypoint map with semantic room labels: $\mathcal{M}_{wp}$
> **Textual Prompt:** You are a robot tasked to locate a person by generating a search plan. The current time is $t$, and your current location is at $\mathbf{w}_t$.
>
> Search query: $q_s, Q'_{db}$.
> Search history: $\mathcal{H}$. Waypoint-to-object assignments $\mathcal{O}_w$.
> Shortest paths from $\mathbf{w}_t$ to $\mathbf{w}_{t+1}$ are $\mathbf{p}_i$.
> Please suggest paths based on the objects at each waypoint to maximize the likelihood of encountering the person along the way, rather than the shortest path. Rank these path variations and assign a score to each path: - path_num, path, score.
>
> The final search plan should be provided in this JSON format: {"waypoints": [{"coordinate": xn, "next_waypoints": [{"waypoint": wp, "objects": […]}, …], "reason": "<reason>"}, ...]}.

Fig. 4: Text and visual prompt of the MLLM Waypoint Planner.

*3) Target Tracking*

The *Target Tracking* module identifies and tracks the target person in the environment. It takes as input an RGB image $\mathbf{x}_{RGB}$, and a text description $q_a$ of the person's appearance. The *Person Tracker*, which runs throughout, detects and tracks individuals using *LDTrack* [6] that we have developed. *LDTrack* leverages diffusion models [44] to capture temporal embeddings of people. When a person is identified, the *Open Detection (OpenDet)* module using G-DINO [45] detects the individual based on their description $q_a$. Each box $b_i$ is associated with a label $c_i$ and confidence score $p(c_i)$, where $c_i \in q_a$. An MLLM is then used to evaluate whether the detected individual matches the search target by comparing $c_i$ with $q_a$. If it matches, a target waypoint $\mathbf{w}_{t+1}$ is passed to the *Navigation* module.

*4) Navigation*

The *Navigation* module converts a target waypoint $\mathbf{w}_{t+1}$ from the *Target Tracking* module or a sequence of waypoints $\mathbf{p}_{t+1}$ from the $MLLM_{WP}$ module into robot velocities $(v, \omega)$ for navigation. The A* algorithm [43] and the TEB planner [46] were used as the global and local planners, respectively. AMCL [47] was used to localize the robot within $\mathcal{M}_{occ}$.

## V. SIMULATED EXPERIMENTS

We conducted extensive simulated experiments to evaluate the performance of our *MLLM-Search* architecture for robotic person search under event-driven scenarios with varying user schedules. Namely, we first conducted an ablation study to investigate the contributions of the design choices of our *MLLM-Search* architecture, and then performed a benchmark comparison study with state-of-the-art (SOTA) person search methods. GPT-4o [28] was used as the MLLM. These experiments were conducted using a Clearpath Jackal robot.

*A. Environments*

Two 3D photorealistic environments from AWS RoboMaker [48] were used in the ROS Gazebo 3D simulator: 1) a hospital environment, and 2) an office environment. The hospital is 25m x 55m in size and includes 11 regions such as patient rooms, intensive care units, and patient wards, with objects such as beds, chairs, and medical equipment, Fig 5(a). The office is 22m x 48m in area and includes 14 regions such as rooms, cubicles, and conference rooms, with objects such as tables, chairs, and TVs, Fig. 5(b). Both environments include dynamic people, whose movements have been modeled using the social force model [49].

*B. Performance Metrics*

We use three metrics to evaluate robot search performance:
1. **Mean Success Rate (SR):** the proportion of searches where the robot successfully locates the target user.
2. **Success weighted by Path Length (SPL):** the efficiency of the search method: $\frac{1}{N}\sum_{i=1}^{N} S_i \frac{l_i}{\max(p_i, l_i)}$, where $N$ is the total number of search trials, $S_i$ represents whether the search was successful, $l_i$ is the shortest path, and $p_i$ is the robot path.
3. **Mean Search Time (ST):** the average time to complete the search and locate the target person across all trials.

*C. Search Scenarios*

For each environment type, we generated three types of scenarios based on people's schedules: 1) **Complete Schedules:** scenarios involving schedules with all information provided; 2) **Shifted Schedules:** scenarios where schedules have been shifted back or forward due to real-time events (i.e., meetings running late or emergencies occurring); 3) **Partial/Incomplete Schedules:** scenarios with partial schedules involving a 1-2 hour gap. Scenarios with incomplete schedules involve larger time gaps of more than 2 hours; and 4) **No Schedules:** scenarios with no prior schedule information available. GPT-4o [28] was used to generate the above scenarios given the waypoint map $\mathcal{M}_{wp}$ and object locations $\mathcal{O}_w$ and $\mathcal{O}_r$. Handcrafted example scenarios of each schedule type were also provided to GPT-4o for in-context learning to generate 10 scenarios for each schedule type.

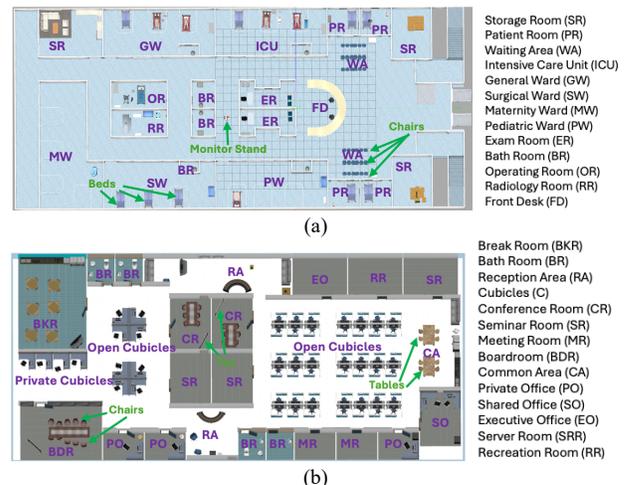

Fig. 5: 3D Gazebo simulation environment of (a) a hospital, and (b) an office.

### D. Ablation Methods

We considered the following for our ablation study: 1) *MLLM-Search*: our proposed method, and 2) *MLLM-Search* **with (w/) Single Stage (SS)**: a variant of *MLLM-Search* using a single MLLM to perform both region and waypoint planning. We also ablated the score variables of the region planner $MLLM_{RP}$ to determine their influence: 3) *MLLM-Search* **without (w/o) Likelihood $s_l$**: a variant with no likelihood score $s_l$; 4) *MLLM-Search* **w/o Proximity $s_p$**: a variant with no proximity score $s_p$; 5) *MLLM-Search* **w/o Recency $s_r$**: a variant with no recency score $s_r$; and 6) *MLLM-Search* **w/o Scores $\Sigma s$** a variant with no scores. Furthermore, we ablated the waypoint planner $MLLM_{WP}$: 7) *MLLM-Search* **w/o SCoT**: a variant with no SCoT prompting.

### E. SOTA Methods

The SOTA methods we compared our *MLLM-Search* method with were: 1) **MDP-based Planner** [20], [21]: this planner selects a region to search based on the expected reward which is determined using transition probabilities between regions and the user location PDF. It was selected as a representative decision-making approach, 2) **HMM-based Planner** [3]: this planner predicts the target person's region by modeling movement as a sequence of hidden states with transition probabilities between regions. It was selected as a representative probabilistic inference approach, and 3) **Random Walk Planner** (RW): RW selects a region to search uniformly at random. It was selected as a naïve baseline. For all methods, GPT-4o is used to generate transition and user location probabilities for each scenario.

### F. Simulation Results

Table I presents the results of both the ablation study and the SOTA comparison. In general, *MLLM-Search* outperformed all methods across all metrics and scenarios. In particular, the ablation study showed that our *MLLM-Search* with two stages of region and waypoint planning performed better than the single stage variant (*MLLM-Search* w/ SS) for both environments. The degradation in performance metrics for the single stage variant is due to the selection of suboptimal waypoints as a result from longer context windows [50]. Namely, the MLLM must simultaneously process more information when combining the planning stages.

Ablating the design choices of the region planner, we observed that removing each score component resulted in performance degradation. The most significant degradation was observed for the *MLLM-Search w/o $\Sigma s$* variant. This variant does not account for the travel distances between regions nor how regions were visited recently, resulting in frequent travel to faraway regions and revisiting the same regions repeatedly. Similarly, *MLLM-Search w/o $s_r$* resulted in the robot frequently revisiting the same regions, leading to longer search times (up to 20.7 min) and up to 50% longer search paths. *MLLM-Search w/o $s_p$* resulted in the robot selecting regions that were farthest away from its current location, leading to inefficient searches as noted by a longer ST of up to 18.1 min as well as a degradation of up to 40% in SR and 48% in SPL. *MLLM-Search w/o $s_l$* prioritizes the closest region rather than the most probable, which resulted in degradations of up to 30% in SR, 36% in SPL, and up to 9.9 min longer ST across all scenarios in both environments.

Ablating the design choice of the waypoint planner, *MLLM-Search w/o SCoT*, resulted in comparable SR and SPL values with *MLLM-Search*, however, up to 22.7 min longer ST. Without SCoT, many of the generated paths were infeasible due to a lack of spatial understanding, thus, requiring more time for GPT-4o to replan.

The comparison results showed that our *MLLM-Search* method outperformed the SOTA planners. For both MDP-based and HMM-based planners, search degradation became more significant as user schedules became less available. The SOTA methods can only perform optimally when a user's past schedule directly matches their actual location, as they rely on historical data to derive transition probabilities (to predict movement patterns) and user location PDFs (to estimate user likelihoods in different regions). However, even with complete schedules, real-time events can cause deviations from the expected schedule. As a result, both MDP- and HMM-based methods only obtained a SR of 0.80-0.90 in the Complete Schedule scenarios across both environments, Table 1. The RW method consistently performed poorly across all scenario and schedule types, often revisiting the same regions or exploring faraway regions.

TABLE I SIMULATION RESULTS

| Environment Type | Hospital | | | | | | | | | | | | Office | | | | | | | | | | | |
|---|---|---|---|---|---|---|---|---|---|---|---|---|---|---|---|---|---|---|---|---|---|---|---|---|
| | Complete Schedule | | | Shifted Schedule | | | Partial / Incomplete Schedule | | | No Schedule | | | Complete Schedule | | | Shifted Schedule | | | Partial / Incomplete Schedule | | | No Schedule | | |
| Method | SR | SPL | ST | SR | SPL | ST | SR | SPL | ST | SR | SPL | ST | SR | SPL | ST | SR | SPL | ST | SR | SPL | ST | SR | SPL | ST |
| *Ablation Methods* | | | | | | | | | | | | | | | | | | | | | | | | |
| *MLLM-Search (our method)* | 1.00 | 0.55 | 8.1 | 1.00 | 0.45 | 10.4 | 0.90 | 0.47 | 12.3 | 0.80 | 0.44 | 15.6 | 1.00 | 0.54 | 9.3 | 1.00 | 0.51 | 11.2 | 1.00 | 0.48 | 13.4 | 0.80 | 0.50 | 14.1 |
| *MLLM-Search w/ SS* | 0.80 | 0.40 | 15.2 | 0.70 | 0.30 | 22.6 | 0.70 | 0.27 | 25.5 | 0.40 | 0.21 | 32.8 | 0.80 | 0.35 | 18.3 | 0.70 | 0.31 | 24.0 | 0.70 | 0.28 | 27.2 | 0.50 | 0.22 | 29.3 |
| *MLLM-Search w/o $s_l$* | 0.80 | 0.44 | 12.5 | 0.70 | 0.33 | 18.3 | 0.60 | 0.35 | 15.2 | 0.50 | 0.31 | 22.4 | 0.90 | 0.38 | 16.6 | 0.80 | 0.36 | 21.1 | 0.70 | 0.37 | 19.3 | 0.50 | 0.32 | 22.9 |
| *MLLM-Search w/o $s_p$* | 0.90 | 0.38 | 18.3 | 0.60 | 0.26 | 26.7 | 0.70 | 0.30 | 23.1 | 0.50 | 0.30 | 24.9 | 0.90 | 0.32 | 20.1 | 0.70 | 0.28 | 29.3 | 0.70 | 0.29 | 25.7 | 0.40 | 0.26 | 27.8 |
| *MLLM-Search w/o $s_r$* | 0.80 | 0.35 | 25.8 | 0.40 | 0.25 | 30.5 | 0.40 | 0.22 | 29.7 | 0.30 | 0.19 | 35.7 | 0.80 | 0.27 | 22.4 | 0.50 | 0.23 | 31.2 | 0.60 | 0.23 | 33.2 | 0.40 | 0.21 | 34.8 |
| *MLLM-Search w/o $\Sigma s$* | 0.60 | 0.25 | 28.1 | 0.30 | 0.15 | 32.9 | 0.40 | 0.16 | 34.7 | 0.30 | 0.17 | 37.1 | 0.60 | 0.25 | 25.7 | 0.50 | 0.22 | 36.0 | 0.50 | 0.15 | 37.7 | 0.30 | 0.18 | 35.3 |
| *MLLM-Search w/o SCoT* | 1.00 | 0.51 | 21.2 | 1.00 | 0.43 | 28.2 | 0.90 | 0.45 | 28.8 | 0.70 | 0.43 | 35.3 | 1.00 | 0.50 | 20.4 | 1.00 | 0.47 | 25.6 | 1.00 | 0.46 | 32.1 | 0.70 | 0.46 | 36.8 |
| *Comparison Methods* | | | | | | | | | | | | | | | | | | | | | | | | |
| MDP-based Planner | 0.80 | 0.31 | 26.3 | 0.40 | 0.22 | 32.5 | 0.40 | 0.26 | 28.4 | 0.30 | 0.15 | 38.0 | 0.80 | 0.24 | 23.4 | 0.50 | 0.21 | 33.1 | 0.60 | 0.22 | 32.7 | 0.40 | 0.20 | 36.5 |
| HMM-based Planner | 0.80 | 0.34 | 18.7 | 0.60 | 0.23 | 28.4 | 0.70 | 0.35 | 22.2 | 0.50 | 0.20 | 30.5 | 0.90 | 0.28 | 20.9 | 0.70 | 0.26 | 31.2 | 0.70 | 0.28 | 25.3 | 0.40 | 0.23 | 32.2 |
| Random Walk | 0.30 | 0.14 | 36.3 | 0.20 | 0.08 | 44.8 | 0.20 | 0.11 | 43.4 | 0.20 | 0.12 | 39.1 | 0.30 | 0.11 | 40.7 | 0.40 | 0.16 | 35.2 | 0.50 | 0.17 | 33.4 | 0.30 | 0.10 | 42.3 |

Under the most challenging scenarios, such as in the No Schedule scenario type, our *MLLM-Search* method outperformed the other SOTA planners with up to 50% improvement in SR, up to 193% improvement in SPL, and achieved ST of up to 22.4 mins faster in both the hospital and office environments. Even without prior user information, our *MLLM-Search* was able to leverage contextual cues from the search scenario to locate the user. For example, in a hospital scenario, the robot was tasked with delivering supplies to a doctor located in the ICU room during an emergency situation. Based on this context, the robot inferred that the doctor was most likely in the Exam Room, ICU, or Operating Room, allowing the robot to efficiently locate the doctor. Similarly, in an office scenario, the robot was tasked with locating a client who arrived for a meeting with the CEO. As visitors do not have schedules in the system, the client's location was unknown. Based on the context of the visit, the robot inferred that the client was most likely in one of three locations: the Reception Area where visitors typically wait, the Conference Room where meetings are often held, or the Executive Office where the CEO might already be meeting the client. This deductive reasoning from contextual information allowed the robot to efficiently locate the user. On the other hand, without prior information, both the MDP and HMM-based methods assumed uniform distributions for user transition probabilities and likelihoods, unable to incorporate semantic understanding of the environment and contextual understanding of the situation.

## VI. REAL-WORLD EXPERIMENTS

We conducted a real-world trial in a 40 m by 43 m multi-room floor of a university building where a food delivery robot was tasked with delivering lunch to a student. The robot consisted of a Jackal robot base with a Velodyne LiDAR and a ZED Mini camera located on a custom platform at a height of 1.2m above the ground, Fig. 6(a). The environment includes the following regions: 1) Conference Room (CR), 2) Lounge (LN), 3) Atrium (AT), 4) Lecture Room (LR), 5) Club Room (CLR), 6) Study Room (SR), and 7) Lab (LB), with various objects such as chairs and tables, Fig. 6(b)-(c). The student was initially located in the AT but moved to the CLR after receiving a message from the 3D printing club regarding a print failure. The waypoint map generated by the *WMG* module is presented in Fig. 7. The search scenario is described as follows:

*Search Query $q_s$:* "Search for an undergraduate student with glasses and wearing jeans. The search start time is 1PM."

*User Schedule $Q_u$:* The student has an exam in two hours at 3PM. This is a partial schedule type due to the time gap.

*Room Scheduling and Student Activity Database $Q_s$:* The scheduling database indicates that the LN is closed due to renovations. Club records indicate that the student is involved with the 3D printing club.

*Results:* The search path taken by the robot, and the path of the student are presented in Fig. 7. The robot first visited the CR as it achieved the highest $s_l$ and $s_p$ scores from $MLLM_{RP}$. CoT reasoning indicated this choice was due to the presence of objects such as chairs and tables, as well as the room being a likely place for students to study for an upcoming exam. After the robot reached the CR, the region scores were updated by the $MLLM_{RP}$, with the AT having the next highest $s_l$ and $s_p$ scores. As a result, the robot navigated to the AT next, with the $MLLM_{WP}$ using our SCoT method to select an obstacle-free route between the tables, providing visibility of seated individuals on both sides of the tables. This demonstrates spatial reasoning as the robot is able to associate objects in the environment with waypoints for path planning. Lastly, the robot visited the CLR, which, despite having a lower $s_l$ score than the SR, had a higher $s_p$ score due to its close proximity to the robot's location resulting in the highest overall score from $MLLM_{RP}$. Overall, *MLLM-Search* achieved a ST of 6.38 min and a SPL of 0.53. A video of this scenario is presented on our YouTube channel at: https://youtu.be/mzP3vcU611Y.

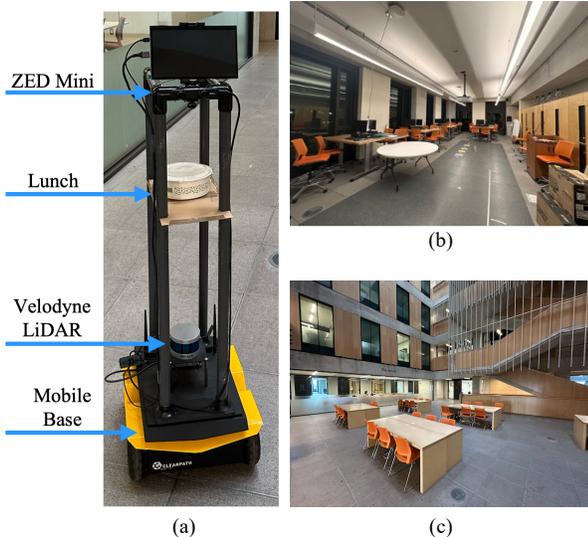

Fig. 6: (a) mobile robot used in the real-world experiments for the food delivery robot scenario; (b-c) images of a floor of a university building.

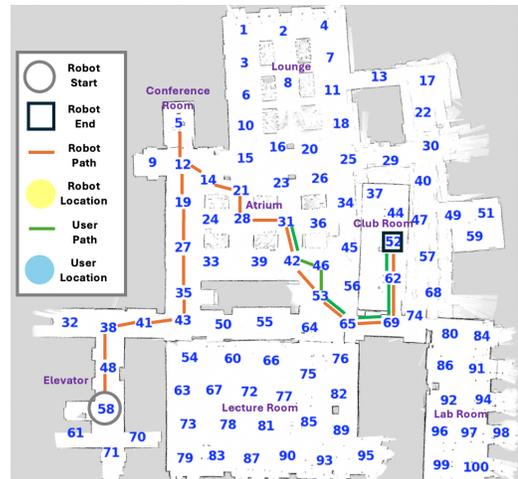

Fig. 7: The waypoint map of the experiment with the search path of the robot (in red), and the path of the student (in green).

## VII. Conclusion

In this paper, we present *MLLM-Search*, a novel person search architecture developed to address the challenge of locating a person under event-driven scenarios with varying user schedules. *MLLM-Search* is the first approach to incorporate MLLMs for search, using a unique visual prompting method that generates a semantically and spatially grounded waypoint map to provide robots with spatial understanding of the global environment. *MLLM-Search* includes a region planner that selects regions based on semantic relevance, and a waypoint planner which considers semantically relevant objects and spatial context using our novel SCoT prompting method in order to plan a robot search path. An extensive ablation study validated the design choices of *MLLM-Search*, while a comparison study with SOTA methods demonstrated that *MLLM-Search* achieves higher search efficiency in 3D photorealistic environments under event-driven scenarios with varying user schedules. A real-world experiment highlight the generalizability of *MLLM-Search* to be applied to a new unseen environment and scenario. Future work will extend *MLLM-Search* to consider human-robot interactions during the search for the robot to obtain additional search evidence or clues.

## References

[1] S. C. Mohamed, S. Rajaratnam, S. T. Hong, and G. Nejat, "Person Finding: An Autonomous Robot Search Method for Finding Multiple Dynamic Users in Human-Centered Environments," IEEE Trans. Autom. Sci. Eng., vol. 17, no. 1, pp. 433–449, Jan. 2020.
[2] S. C. Mohamed, A. Fung, and G. Nejat, "A Multirobot Person Search System for Finding Multiple Dynamic Users in Human-Centered Environments," IEEE Trans. Cybern., vol. 53, no. 1, pp. 628–640, Jan. 2023.
[3] P. Bovbel and G. Nejat, "Casper: An Assistive Kitchen Robot to Promote Aging in Place1," J. Med. Devices, vol. 8, no. 030945, 2014.
[4] A. Bayoumi, P. Karkowski, and M. Bennewitz, "Speeding up person finding using hidden Markov models," Robot. Auton. Syst., vol. 115, pp. 40–48, 2019.
[5] P. Elinas, J. Hoey, and J. J. Little, "HOMER: Human Oriented MEssenger Robot," AAAI Symp. Human Inter. Auton. Syst., pp. 45-51, 2003.
[6] A. Fung, B. Benhabib, and G. Nejat, "LDTrack: Dynamic People Tracking by Service Robots using Diffusion Models." arXiv, 2024.
[7] X. Lin, R. Lu, D. Kwan, and X. Shen, "REACT: An RFID-based privacy-preserving children tracking scheme for large amusement parks," Comput. Netw., vol. 54, no. 15, pp. 2744–2755, Oct. 2010.
[8] D. Dworakowski, A. Fung, and G. Nejat, "Robots Understanding Contextual Information in Human-Centered Environments Using Weakly Supervised Mask Data Distillation," Int. J. Comput. Vis., vol. 131, no. 2, pp. 407–430, Feb. 2023.
[9] A. Fung, L. Y. Wang, K. Zhang, G. Nejat, and B. Benhabib, "Using Deep Learning to Find Victims in Unknown Cluttered Urban Search and Rescue Environments," Curr. Robot. Rep., vol. 1, no. 3, pp. 105–115, Sep. 2020.
[10] A. H. Tan, F. P. Bejarano, Y. Zhu, R. Ren, and G. Nejat, "Deep Reinforcement Learning for Decentralized Multi-Robot Exploration With Macro Actions," IEEE Robot. Autom. Lett., vol. 8, no. 1, pp. 272–279, Jan. 2023.
[11] H. Wang, A. Tan, and G. Nejat, "NavFormer: A Transformer Architecture for Robot Target-Driven Navigation in Unknown and Dynamic Environments," IEEE Robot. Autom. Lett., vol. PP, pp. 1–8, Aug. 2024.
[12] D. McColl, W.-Y. G. Louie, and G. Nejat, "Brian 2.1: A socially assistive robot for the elderly and cognitively impaired," IEEE Robot. Autom. Mag., vol. 20, no. 1, pp. 74–83, Mar. 2013.
[13] M. Montemerlo, et al., "Experiences with a mobile robotic guide for the elderly," Int. Conf. Artif. Intelli., USA, Jul. 2002, pp. 587–592.
[14] G. D. Tipaldi and K. O. Arras, "I want my coffee hot! Learning to find people under spatio-temporal constraints," IEEE Int. Conf. Robot. Autom., Shanghai, China, May 2011, pp. 1217–1222.
[15] A. Fung, B. Benhabib, and G. Nejat, "Robots Autonomously Detecting People: A Multimodal Deep Contrastive Learning Method Robust to Intraclass Variations," Robot. Autom. Lett., vol. 8, no. 6, pp.3550–3557, 2023.
[16] M. K. Hasan, A. Hoque, and T. Szecsi, "Application of a Plug-and-Play Guidance Module for Hospital Robots," 2010.
[17] M. Volkhardt and H.-M. Gross, "Finding people in home environments with a mobile robot," Conf. Mobile Robots, Spain, Sep. 2013, pp. 282–287.
[18] H. Lau, S. Huang, and G. Dissanayake, "Optimal search for multiple targets in a built environment," Int. Conf. Intell. Robots Syst., Edmonton, Alta., Canada, 2005, pp. 3740–3745.
[19] S. Lin and G. Nejat, "Robot Evidence Based Search for a Dynamic User in an Indoor Environment," Mechanisms Robot. Conf., Quebec, Canada, 2018.
[20] S. A. Mehdi and K. Berns, "Behavior-based search of human by an autonomous indoor mobile robot in simulation," Univers. Access Inf. Soc., vol. 13, no. 1, pp. 45–58, Mar. 2014.
[21] M. Schwenk, T. Vaquero, G. Nejat, and K. Arras, "Schedule-Based Robotic Search for Multiple Residents in a Retirement Home Environment," Proc. AAAI Conf. Artif. Intell., vol. 28, no. 1, 2014.
[22] B. Yu, H. Kasaei, and M. Cao, "L3MVN: Leveraging Large Language Models for Visual Target Navigation," IEEE Int. Conf. Intell. Robot. Syst., Oct. 2023, pp. 3554–3560.
[23] Y. Kuang, H. Lin, and M. Jiang, "OpenFMNav: Towards Open-Set Zero-Shot Object Navigation via Vision-Language Foundation Models." arXiv, 2024.
[24] D. Shah, B. Osinski, B. Ichter, and S. Levine, "LM-Nav: Robotic Navigation with Large Pre-Trained Models of Language, Vision, and Action." arXiv, 2022.
[25] G. Zhou, Y. Hong, and Q. Wu, "NavGPT: Explicit Reasoning in Vision-and-Language Navigation with Large Language Models." arXiv, 2023.
[26] T. B. Brown et al., "Language Models are Few-Shot Learners." arXiv, 2020.
[27] H. Liu, R. Ning, Z. Teng, J. Liu, Q. Zhou, and Y. Zhang, "Evaluating the Logical Reasoning Ability of ChatGPT and GPT-4." arXiv, 2023.
[28] OpenAI, "GPT-4o," 2024.
[29] A.-C. Cheng et al., "SpatialRGPT: Grounded Spatial Reasoning in Vision Language Model." arXiv, 2024.
[30] A. Hao Tan, A. Fung, H. Wang, and G. Nejat, "Mobile Robot Navigation Using Hand-Drawn Maps: A Vision Language Model Approach." arXiv, 2024.
[31] X. Lei, Z. Yang, X. Chen, P. Li, and Y. Liu, "Scaffolding Coordinates to Promote Vision-Language Coordination in Large Multi-Modal Models." arXiv, 2024.
[32] J. Yang, H. Zhang, F. Li, X. Zou, C. Li, and J. Gao, "Set-of-Mark Prompting Unleashes Extraordinary Visual Grounding in GPT-4V." arXiv, 2023.
[33] A.-C. Cheng et al., "SpatialRGPT: Grounded Spatial Reasoning in Vision Language Model." arXiv, 2024.
[34] S. Nasiriany et al., "PIVOT: Iterative Visual Prompting Elicits Actionable Knowledge for VLMs." arXiv, 2024.
[35] T. Ren et al., "Grounded SAM: Assembling Open-World Models for Diverse Visual Tasks." arXiv, 2024.
[36] D. S. Chaplot, D. P. Gandhi, A. Gupta, and R. R. Salakhutdinov, "Object Goal Navigation using Goal-Oriented Semantic Exploration," in Adv. Neural Inf. Process. Syst., 2020, vol. 33, pp. 4247–4258.
[37] J. Rebello, A. Fung, and S. L. Waslander, "AC/DCC : Accurate Calibration of Dynamic Camera Clusters for Visual SLAM," in IEEE Int. Conf. Robot. Automat., May 2020, pp. 6035–6041.
[38] A. Hao Tan, A. Al-Shanoon, H. Lang, and M. El-Gindy, "Mobile Robot Regulation With Image Based Visual Servoing."
[39] G. Grisetti, C. Stachniss, and W. Burgard, "Improved Techniques for Grid Mapping With Rao-Blackwellized Particle Filters," IEEE Trans. Robot., vol. 23, no. 1, pp. 34–46, Feb. 2007.
[40] J. E. Bresenham, "Algorithm for computer control of a digital plotter," IBM Syst. J., vol. 4, no. 1, pp. 25–30, 1965.
[41] P. Lewis et al., "Retrieval-Augmented Generation for Knowledge-Intensive NLP Tasks." arXiv, 2021.
[42] J. Wei et al., "Chain-of-Thought Prompting Elicits Reasoning in Large Language Models." arXiv, 2023.
[43] P. E. Hart, N. J. Nilsson, and B. Raphael, "A Formal Basis for the Heuristic Determination of Minimum Cost Paths," IEEE Trans. Syst. Sci. Cybern., vol. 4, no. 2, pp. 100–107, Jul. 1968.
[44] A. H. Tan, S. Narasimhan, and G. Nejat, "4CNet: A Confidence-Aware, Contrastive, Conditional, Consistency Model for Robot Map Prediction in Multi-Robot Environments." arXiv, 2024.
[45] S. Liu et al., "Grounding DINO: Marrying DINO with Grounded Pre-Training for Open-Set Object Detection." arXiv, Mar. 20, 2023.
[46] C. Rösmann, W. Feiten, T. Woesch, F. Hoffmann, and T. Bertram, "Trajectory modification considering dynamic constraints of autonomous robots," Jan. 2012, pp. 1–6.
[47] F. Dellaert, D. Fox, W. Burgard, and S. Thrun, "Monte Carlo localization for mobile robots," Int. Conf. Robot. Automat., 1999, vol. 2, pp. 1322–1328 vol.2.
[48] "AWS robomaker: Amazon cloud robotics platform."
[49] D. Helbing and P. Molnar, "Social Force Model for Pedestrian Dynamics," Phys. Rev. E, vol. 51, no. 5, pp. 4282–4286, May 1995.
[50] N. F. Liu et al., "Lost in the Middle: How Language Models Use Long Contexts." arXiv, 2023.